\documentclass[letterpaper, 10pt, conference]{ieeeconf}  

\usepackage{graphicx}
\usepackage{float}
\usepackage[utf8]{inputenc}
\usepackage[T1]{fontenc}
\usepackage{amsmath}

\IEEEoverridecommandlockouts                              % This command is only needed if 
\title{\LARGE \bf
	A Decentralized Interactive Architecture for Aerial and Ground Mobile Robots Cooperation
}

\author{El Houssein Chouaib Harik$^{1}$,  François Guérin$^{2}$, Frédéric Guinand$^{1}$,\\ Jean-François Brethé$^2$, Hervé Pelvillain$^3$% <-this % stops a space
	\thanks{$^{1}$LITIS, University of Le Havre. 25 Rue Philippe Lebon, 76600 Le Havre Cedex (France) -  
		(el-houssein-chouaib.harik,frederic.guinand)@univ-lehavre.fr}% <-this % stops a space
	\thanks{$^{2}$GREAH, University of Le Havre. 75, Rue Bellot, 76058 Le Havre Cedex (France) - 
		(francois.guerin,jean-francois.brethe)@univ-lehavre.fr}%
	\thanks{$^{3}$IUT GEII, University of Le Havre. Rue Boris Vian, 76610 Le Havre Cedex (France) -
		herve.pelvillain@univ-lehavre.fr}%
}

\begin{document}

	\maketitle
	\thispagestyle{empty}
	\pagestyle{empty}

\begin{abstract}
\textit{This paper presents a novel decentralized interactive architecture for aerial and ground mobile robots cooperation. The aerial mobile robot is used to provide a global coverage during an area inspection, while the ground mobile robot is used to provide a local coverage of ground features. We include a human-in-the-loop to provide waypoints for the ground mobile robot to progress safely in the inspected area. The aerial mobile robot follows continuously the ground mobile robot in order to always keep it in its coverage view.}
\end{abstract}
	
	% % % % % % % % % % % % % % % % % % % % % % % % % % % % % % % % % % % % % % % % % % % % %
\section{INTRODUCTION}
Unmanned Aerial and Ground Vehicles (UAVs, UGVs) cooperation attracts increasingly the attention of researchers, essentially for the complementary skills provided by each type to overcome the specific limitations of the others. UAVs provide a global coverage and faster velocities, and UGVs provide a higher payload and stronger calculation capabilities, adding a local coverage for the unseen areas from an aerial perspective. In addition, deploying both types together ensure faster and more reliable results within a shorter frame of time compared to the deployment of a single type of mobile robots. 
	
Air-Ground-Cooperation (AGC) in mobile robots systems can envisage a large panel of applications. Our focus in this paper is on Surveillance or Inspection (SI) missions. The authors in \cite{ridley2002tracking}, \cite{grocholsky2005information},\cite{grocholsky2006cooperative} use a Decentralized Data Fusion (DDF) technique \cite{manyika1995data} for exploring a specific area. The task of both air and ground based nodes is to make observations of terrain features and identify moving or stationary targets.

The authors in \cite{chaimowicz2004framework} use AGC for target searching missions. Their testbed is composed of three layers: a high altitude UAV (a planner) to determine the motion of the mobile robots based on the desired goals, medium altitude UAVs (Blimps) to track the group of Unmanned Ground Vehicles (UGVs) and try to maintain them in their field of view, and the UGVs that navigates for exploration or other tasks. Another interesting work can be found in \cite{chaimowicz2007aerial}, where the authors present a system that allow to deploy a significant number of ground mobile robots monitored by aerial mobile robots (aerial shepherds) for target searching.

An urban environment surveillance application can be found in \cite{hsieh2007adaptive}. The authors of this work cover a large set of technological topics from mapping to communication and collaborative navigation using multiple aerial and ground mobile robots.

The authors in \cite{tanner2007decentralized} also use a group of UAVs to provide an aerial coverage to a group of UGVs used to clear a path through that area. 

We can find in \cite{tanner2006cooperation} a surveillance scheme including a group of six UGVs and four UAVs. An interagent cohesion and separation and a velocity synchronization scheme were combined into a control and communication strategy for mobile robots navigation. Another work of the same authors \cite{tanner2007switched} present an AGC scheme for target detection mission. The size of the navigation area is related to the number of the deployed mobile robots. A set of grid points is defined for UGVs. An interagent potential force based approach is used for the UGVs navigation towards the grid points. Once the UGVs reached their grid points the aerial mobile robots start scanning the field. 

The authors in \cite{stentz2002integrated,stentz2003real} use a UAV as a remote sensor that flies ahead the UGV to provide geo-referenced 3D geometry, in order to help the UGV for navigating in the area avoiding obstacles.

In the previous related works, a first flight of the drone over the inspected area is always performed before the beginning of the mission. A traversability map is then processed to provide a trajectory to the UGV. The main contribution of our work consists in providing a real-time navigation scheme: the UAV flies over the area to provide a global coverage, and to assist a UGV in real-time to navigate safely avoiding obstacles. To provide the real-time navigation scheme without the need to have a per-processed map, a human operator is used to provide waypoints by monitoring the global and local coverages, making the architecture interactive, and applicable to real scenarios. Indeed, for numerous reasons, monitoring, surveillance and inspection of industrial sites belonging to Seveso category (highly critical sites) cannot be performed without a constant and careful attention of human experts.

The paper is organized as follows: In section \ref{pb_st} we introduce the problem statement and define the objectives of our work. Section \ref{uav_ugv} presents the UGV and UAV controllers design. We discuss in section \ref{ugv_res} and \ref{uav_res} the simulation and experimental results, and we conclude in section \ref{conclusion} the present work and discuss future perspectives.

\section{Problem statement}\label{pb_st}
 
\begin{figure}[H]
	\centering 
	\includegraphics[width = 230px , height =140px]{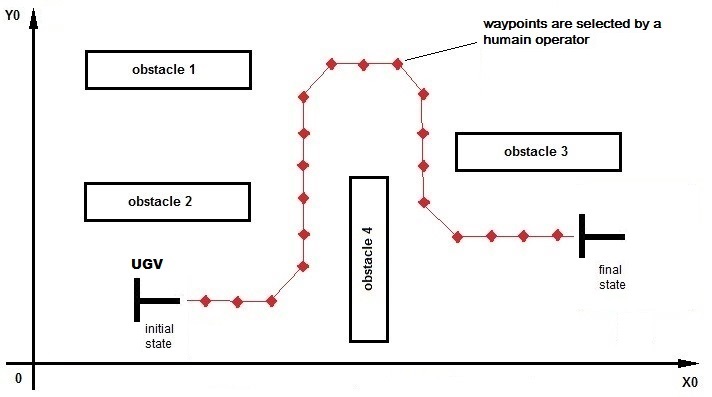} 
	\caption{Problem statement}
	\label{first_scen} 
\end{figure}

\begin{figure}[H] 
	\centering
	\includegraphics[width = 199px , height = 140px]{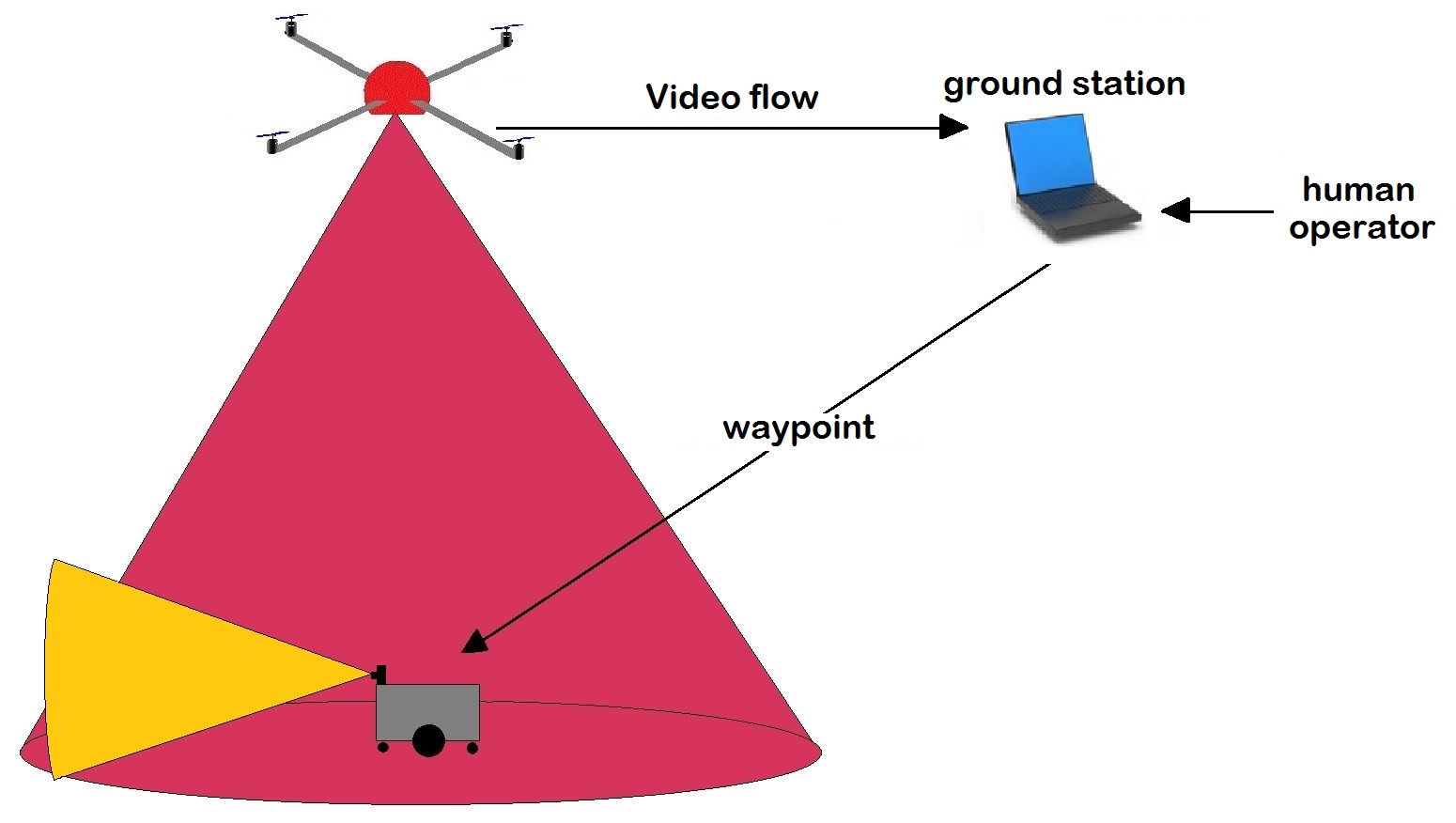} 
	\caption{Overall architecture}
	\label{arch} 
\end{figure}

A UAV is used to monitor a given area where a UGV is present to perform the inspection task. The images taken from the camera mounted on the UAV are sent continuously to the ground station. The images are processed in real time to provide localization information to the UGV to navigate in the area through the on-the-go waypoints selected by a human operator (Figures \ref{first_scen} and \ref{arch}).

To successfully inspect a given area within a reasonable frame of time, global and local coverages must be acquired simultaneously, which can be possible if we deploy both types of mobile robots (aerial and ground) simultaneously. In this direction, our architecture is composed of a UAV equipped with a down-facing camera, and a UGV equipped with a horizontal-facing camera. Both video flows are sent continuously to a ground station running our developed Human-Machine-Interface (HMI). A human operator supervises the video flows, and selects progressively navigation waypoints for the UGV. As the UGV navigates through the given waypoints, the UAV follows continuously the UGV in order to keep it in its coverage view (the center of the image plan) using visual servoing.  
	
\section{UGV and UAV controllers design} \label{uav_ugv}

\subsection{Hardware configuration}
We consider in this work a non-holonomic, unicycle-like, Wheeled Mobile Robot (WMR) that has been designed in the GREAH laboratory of the University of Le Havre (France) \cite{Guerin2013}. The UAV is a quadcopter (Phantom 2 Vision) developed by DJI \cite{dji} with a sufficient payload to carry an open platform device (android phone) that is used to take the video flow and send it to the ground station continuously via Wifi (IEEE 802.11). The received flow is processed to extract the position and orientation (pose estimation) of the UGV relative to the selected waypoint.

\subsection{Pose estimation of the UGV}\label{pose}
To successfully navigate in a given area, the relative pose of the UGV should be known (orientation and location). The authors of AGC applications (\cite{choidead}, \cite{chaimowicz2004experiments}, \cite{vidal2002probabilistic,kim2001hierarchical}) used a single colored marker to define the relative location of the UGV to the UAV, and a digital compass to communicate the orientation which generally gives inaccurate measurements (digital compass is sensible to electromagnetic field variations). To overcome this issue, we added a second marker to also estimate its orientation.

\begin{figure}[H]
	\centering
	\includegraphics[width = 180px , height =120px]{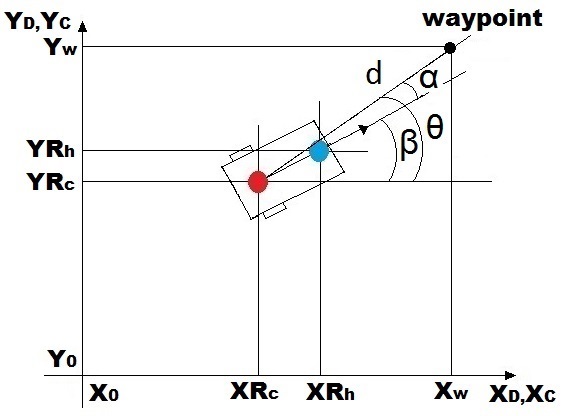} 
	\caption{The UGV in the image plan}
	\label{robot_track_} 
\end{figure}

We consider \(X_{c}\) and  \(Y_{c}\) the axis of the image (camera) taken from the UAV (Figure \ref{robot_track_}). In order to locate the UGV, we used a color tracking algorithm to extract the position of colored markers: \(R_c\) (red marker) fixed on the center of the driving wheels, and \(R_h\) (blue marker) fixed on the head of the UGV. The position (in pixels) of \(R_c\) and \(R_h\) along \(X_{c}\) and  \(Y_{c}\) axis are respectively: \(Y_{Rc}\), \(X_{Rc}\) and \(Y_{Rh}\), \(X_{Rh}\). We assume that the center of the camera (center of the image) is the center of the UAV, thus we assume that the camera frame (\(X_{c}\),\(Y_{c}\)) and the UAV inertial frame (\(X_{D}\), \(Y_{D}\)) are superimposed.

The orientation \(\alpha\) of the UGV to a given waypoint \(w\), can be computed as \(\alpha=\theta-\beta\) with:

\begin{equation}
\theta= atan(\frac{Y_w-Y_{Rc}}{X_w-X_{Rc}}) \hspace{5 mm} \beta= atan(\frac{Y_{Rh}-Y_{Rc}}{X_{Rh}-X_{Rc}})
\label{beta}
\end{equation}    

Note that by using the colored markers on the UGV, we can extract its orientation without knowing the orientations of the UAV and the UGV in the world frame. The distance \(d\) that separates the UGV and the waypoint \(w\) has to be expressed in the world frame. It can be written as follows:

\begin{equation}
d=\sqrt{(x_w-x_{Rc})^2+(y_w-y_{Rc})^2}
\end{equation}

Where \(x_w, y_w, x_{Rc}, y_{Rc} \) represent (in meter) the coordinates of the \(R_c\) marker and the waypoint in the UAV or camera frame previously supposed to be the same. They can be computed using the classical Pinhole model of the camera \cite{francoisthesis}. The example is given for the waypoint: 

\begin{equation} \left[ \begin{array}{c} X_{w}.z \\ Y_{w}.z \\ z \end{array}  
\right] = \begin{bmatrix}
G_x & 0 & 0 & z.X_0 \\
0 & G_y & 0 & z.Y_0 \\
0 & 0 & 1 & 0     
\end{bmatrix} \left[ \begin{array}{c} x_w \\ y_w \\ z \\ 1 \end{array} 
\right]
\label{pinhole}
\end{equation}

\hspace{1.5cm}\(G_x =\frac{f}{\Delta_x} \hspace{1cm} G_y =\frac{f}{\Delta_y}\)\\\\
\noindent Where:\\
\(f\): Focal distance - \(\Delta_x, \Delta_y\): Dimensions of a pixel.\\
\(X_0, Y_0\): Projection (in pixels) of the camera optical center in the image plan. They are assumed to be null.\\
\(X_w, Y_w\): Projection (in pixels) of the waypoint in the image plan.\\
We note as well that using the pinhole model we can estimate the altitude of the UAV using the distance between the two colored markers (red and blue).

\subsection{Design of the UGVs controller} \label{kinematic_controller}
\begin{figure}[H] 
	\centering
	\includegraphics[scale=0.32]{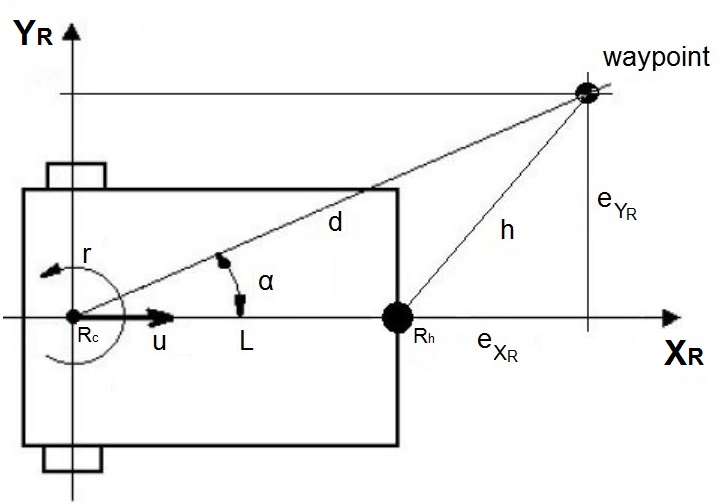} 
	\caption{Design of the UGV controller}
	\label{leader} 
\end{figure}

The UGV receives \(d\) and \(\alpha\) via a radio module from the ground station. The goal is to use these information to generate \(u\) and \(r\) that respectively represent the longitudinal and angular velocities of the UGV (Figure \ref{leader}). The control objective is to regulate the distance \(d\) and the steering angle \(\alpha\) as follows:

\begin{equation}
\lim \limits_{t \to \infty} d(t) = L  \hspace{5 mm} \lim \limits_{t \to \infty} \alpha(t) = 0
\label{leader_obj}
\end{equation}

Let us consider the following distance errors on \(X_R\) and \(Y_R\) axes respectively:

\begin{equation}
\begin{aligned}
e_{X_R} = d.cos(\alpha)-L\\
e_{Y_R} = d.sin(\alpha)
\end{aligned}
\label{dist_err_lead}
\end{equation}

The time derivative of the distance errors (\ref{dist_err_lead}), which depends on both the UGV's velocities (\(u,r\)) and the target's velocities along the \(X_R\) and \(Y_R\) axes, denoted as \(TV_{X_R}\) and \(TV_{Y_R}\), are given by:
\begin{equation}
\begin{aligned}
\dot{e}_{X_R} = \dot{d}.cos(\alpha)-d.\dot{\alpha}.sin(\alpha)= TV_{X_R}-u\\
\dot{e}_{Y_R} = \dot{d}.sin(\alpha)+d.\dot{\alpha}.cos(\alpha)= TV_{Y_R}-rL
\end{aligned}
\label{der_dist_err_lead}
\end{equation}

Where \(rL\) corresponds to the linear velocity of \(R_h\) (blue marker).\\ 

\(TV_{X_R}\) and \(TV_{Y_R}\) are null since the target of the UGV is a waypoint. We propose then the following error dynamic equations:

\begin{equation}
\begin{aligned}
\dot{e}_{X_R}=-K.e_{X_R}=-u  \hspace{5 mm}
\dot{e}_{Y_R}=-K.e_{Y_R}=-r.L
\end{aligned}
\label{cmd_law_lead1}
\end{equation}

Where \(K>0\) is the proportional gain of the UGV's controller. It is chosen to get a first order dynamic behavior for the control objective:

\begin{equation}
\begin{aligned}
\dot{e}_{X_R}+K.{e}_{X_R}=0    \hspace{5 mm}
\dot{e}_{Y_R}+K.{e}_{Y_R}=0 
\end{aligned}
\label{first_ord}
\end{equation}

The numerical value given to \(K\) allows to define the decreasing speed of the distance errors \({e}_{X_R}\) and \({e}_{Y_R}\). Note that for large values of the distance \(d\), a saturation is implemented to consider the limitations of the electrical drives that control the UGV's left and right wheels.

Starting from (\ref{cmd_law_lead1}) we propose the following non-linear kinematic controller: 

\begin{equation}
\begin{aligned}
u = K.(d.cos(\alpha)-L) \\
r = K.(d.sin(\alpha))/L
\end{aligned}
\label{cmd_law_lead}
\end{equation}

\subsection{Design of the UAV controller} 
To hover the UAV over the UGV during its waypoints navigation, we need to control the UAV movements along \(X_D\) and \(Y_D\) axis (Figure \ref{robot_track_}) since the UGV moves on a 2D plan (assumed to be flat). We need to develop a navigation controller that takes as input the location of the UGV (\(R_c\)), and generates the necessary pitch and roll angles to keep the UGV at the center of the image plan (visual servoing).

We consider four inputs \(\phi_d, \theta_d, \psi_d\) and \(z_d\). They represent respectively the desired roll, pitch, yaw angles, and the desired altitude. Note that the UAV attitude (roll, pitch, yaw, altitude) is controlled by the internal autopilot. We suppose that the UAV flies at a fixed altitude, and a fixed yaw: \(z_d= z\) and \(\psi_d = 0\). \(z\) is chosen to have a sufficient coverage (\ref{pinhole}) of the inspected area (\(z_{min} < z \)), and to keep the track of the UGV (\(z < z_{max}\)). \(z_{min}\) and \(z_{max}\) can be defined experimentally. 

To design the UAV controller, we used the dynamic model of a quadcopter \cite{bouabdallah2007full}. Starting from this dynamic model, the movements along \(X_D\) and \(Y_D\) axis can be described as follows:

\begin{equation}
\begin{aligned}
\ddot{x}_D = (cos\phi_d sin\theta_d  cos\psi_d+sin\phi_d sin\psi_d)\frac{1}{m}U_1\\
\ddot{y}_D = (cos\phi_d sin\theta_d  sin\psi_d-sin\phi_d cos\psi_d)\frac{1}{m}U_1
\end{aligned}
\label{interests}
\end{equation}

\(\ddot{x}_D\) and \(\ddot{y}_D\) correspond to the acceleration along \(X_D\) and \(Y_D\) axis respectively. (\ref{interests}) can be as:

\begin{equation} 
\begin{aligned}
\ddot{x} = u_{X_{D}} \frac{1}{m}U_1\\
\ddot{y} = u_{Y_{D}} \frac{1}{m}U_1
\end{aligned}
\end{equation}

With:

\begin{equation}
\begin{aligned}
u_{X_{D}} = (cos\phi_d sin\theta_d cos\psi_d+sin\phi_d sin\psi_d)\\
u_{Y_{D}} = (cos\phi_d sin\theta_d sin\psi_d-sin\phi_d cos\psi_d)
\end{aligned}
\label{lineare}
\end{equation}

where \(u_{X_{D}}, u_{Y_{D}}\) represent the orientations of the total thrust (\(U_1\)) responsible for linear motions of the quadcopter along \(X_D\) and \(Y_D\) axis. \(m\) is the UAVs weight.

To allow the UAV to keep the UGV in its coverage view (the center of the image plan), let us consider the following distance errors (Figure \ref{robot_track_}):

\begin{equation}
\begin{aligned}
e_{X_{D}}= X_{Rc}-X_0\hspace{5 mm}
e_{Y_{D}}=Y_{Rc}-Y_0
\end{aligned}
\label{err}
\end{equation}

Note that \(X_0, Y_0\) are assumed to be null. The control objective is to regulate the distance errors as follows:

\begin{equation}
\begin{aligned}
\lim \limits_{t \to \infty} e_{X_{D}}(t) = 0 \hspace{5 mm}
\lim \limits_{t \to \infty} e_{Y_{D}}(t) = 0
\end{aligned}
\label{uav_obj}
\end{equation} 

To fulfill this control objective, we propose the following error dynamic equations:

\begin{equation}
\begin{aligned}
\ddot{e}_{X_{D}}= -K_{1_{XD}}e_{X_{D}} - K_{2_{XD}}\dot{e}_{X_{D}} = TA_{XD}- \ddot{x}_D \\
\ddot{e}_{Y_{D}}= -K_{1_{YD}}e_{Y_{D}} - K_{2_{YD}}\dot{e}_{Y_{D}} = TA_{YD}- \ddot{y}_D
\end{aligned}
\label{dyn_err}
\end{equation}

Where \(K_{(1,2),(XD,YD)}\) are positive gains.  They are chosen to get a second order dynamic behavior without damping and oscillations for the control objective:

\begin{equation}
\begin{aligned}
\ddot{e}_{X_{D}}+K_{2_{XD}}\dot{e}_{X_{D}}+K_{1_{XD}}e_{X_{D}} = 0 \\
\ddot{e}_{Y_{D}}+K_{2_{YD}}\dot{e}_{Y_{D}}+K_{1_{YD}}e_{Y_{D}} = 0
\end{aligned}
\label{ctrl_obj}
\end{equation}

\(TA_{XD}\) and \(TA_{YD}\) are the accelerations of the UGV along \(X_D, Y_D\) axis. They are neglected compared to \(\ddot{x}_D\) and \(\ddot{y}_D\) since the UAV moves much faster than the UGV. Starting from (\ref{dyn_err}), we propose the following controller:

\begin{equation} 
\begin{aligned}
u_{X_{D}} = \frac{m}{U_1}(K_1 e_{X_{D}} + K_2 \dot{e}_{xD})\\
u_{Y_{D}} = \frac{m}{U_1}((K_1 e_{Y_{D}} + K_2 \dot{e}_{yD})
\end{aligned}
\label{PD_control}
\end{equation}

To find the desired roll and pitch angles needed to track the UGV, we combine (\ref{PD_control}) in (\ref{lineare}) taking a fixed yaw angle \(\psi=0\), the desired angles are then:

\begin{equation}  
\begin{aligned}
\phi_d = asin(-u_{Y_{D}})\\
\theta_d = asin(\frac{u_{X_{D}}}{cos\phi_d})
\end{aligned}
\label{desired_angles}
\end{equation}

We note that the steady state error will be null (when the UGV stops) since it is a position control.

\subsection{DES algorithm}
Furthermore, the Double Exponential Smoothing (DES) algorithm \cite{laviola2003experiment} has been implemented to smoothen the uncertain measurements given by the vision system. The DES algorithm runs approximately 135 times faster with equivalent prediction performances and simpler implementations compared to a Kalman filter.
The DES algorithm  at time instant \(n.T_e\), where \(T_e\) is the sampling period and \(n\) is the discrete-time index, implemented for the distance \(d\), the steering angle \(\alpha\), and the projection of the red marker \(X_{Rc}\), \(Y_{Rc}\) is given as follows:

\begin{equation}
S_{m_n}=\gamma_m.m_n+(1-\gamma_m).(S_{m_{n-1}}+b_{m_{n-1}})
\label{smooth_d}
\end{equation}
\begin{equation}
b_{m_n}=\lambda_m.(S_{m_{n}}-S_{m_{n-1}})+(1-\lambda_m).b_{m_{n-1}}
\label{trend_d}
\end{equation}

\noindent Where:\\
\(m_n\) is the value of (\(d\),\(\alpha\),\(X_{Rc}\), \(Y_{Rc}\)) at \(n\)th sample instant.\\
\(S_{m_n}\) is the smoothed values of  (\(d\),\(\alpha\),\(X_{Rc}\), \(Y_{Rc}\)).\\
\(b_{m_n}\)is the trend value of (\(d\),\(\alpha\),\(X_{Rc}\), \(Y_{Rc}\)).\\

Equation (\ref{smooth_d}) smooths the value of the sequence of measurements by taking into account the trend, whilst (\ref{trend_d}) smooths and updates the trend.

The initial values given to \(S_{m_n}, b_{m_n}\) are:

\begin{equation}
\begin{aligned}
S_{m_1}= m_1 \hspace{5 mm}
b_{m_1}= m_2-m_1 
\end{aligned}
\end{equation}

Usually, \(\gamma\) (\(0 \leq \gamma \leq 1\)) is called the data smoothing factor and \(\lambda\) (\(0 \leq \lambda \leq 1\)) is called the trend smoothing factor. A compromise has to be found for the values of \(\gamma\) and \(\lambda\). High values make the DES algorithm follow the trend more accurately whilst small values make it generate smoother results. The smoothed values are used instead of the direct noisy measurements in the proposed controllers. DES algorithm has been used successfully in a previous work \cite{Guerin2013} for vision based target tracking.

\section{UGV results}\label{ugv_res}
We will present in this section simulation and experimental results concerning only the UGV, the UAV is piloted by a professional.

\subsection{Simulation results}
We simulated a real-like scheme to navigate in a given area avoiding obstacles. For the UGV we have used the dynamic model fully described in \cite{Guerin2013} and \cite{francoisthesis}. Simulations have been carried out with Matlab-Simulink software (sampling period: 10 ms). For the UGV non-linear kinematic controller we used the following parameters: \(L=15cm , K=0.1\).

Figure \ref{simulation} shows that the UGV is able to follow the given waypoints, which validates the proposed controller (\ref{cmd_law_lead}). The results concerning the control objective (\ref{leader_obj}) have been checked experimentally.

\subsection{Experimental results}

\begin{figure}[H]
	\centering
	\includegraphics[width = 220px , height = 95px]{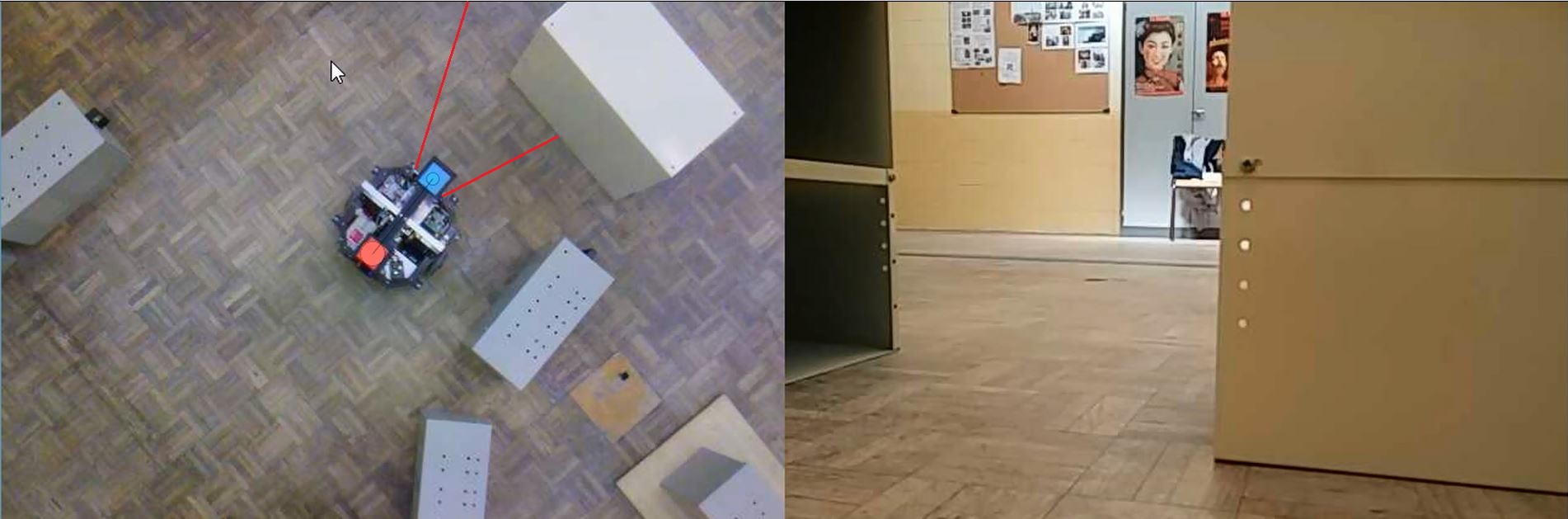} 
	\caption{Aerial and ground perception on the HMI. Left: The global view from the UAV. Right: the local view from the UGV.}
	\label{perception} 
\end{figure}

We created a user interface running on the ground station using Processing \cite{Processing}. The human operator can supervise in real time both video flows received from UAV and UGV (Figure \ref{perception}).

The human operator can select waypoints in the image (global view side) by a simple click. As explained in section (\ref{pose}), the software extracts and sends the distance \(d\) and the steering angle \(\alpha\) to the UGV via ZigBee module (IEEE 802.15.4) at a frequency of 50Hz (real time interrupt). 

The experiments were carried out with a smartphone mounted under the quadcopter. For safety reasons, the UAV was piloted by a professional, but we carried out as well other experiments without a pilot using an AR Drone 2.0. Videos can be found in \cite{colortrack} for the experiment with a pilot, and \cite{experiment} for the newer version without a pilot .  

\begin{figure}[H] 
	\centering
	\includegraphics[width = 220px , height = 95px]{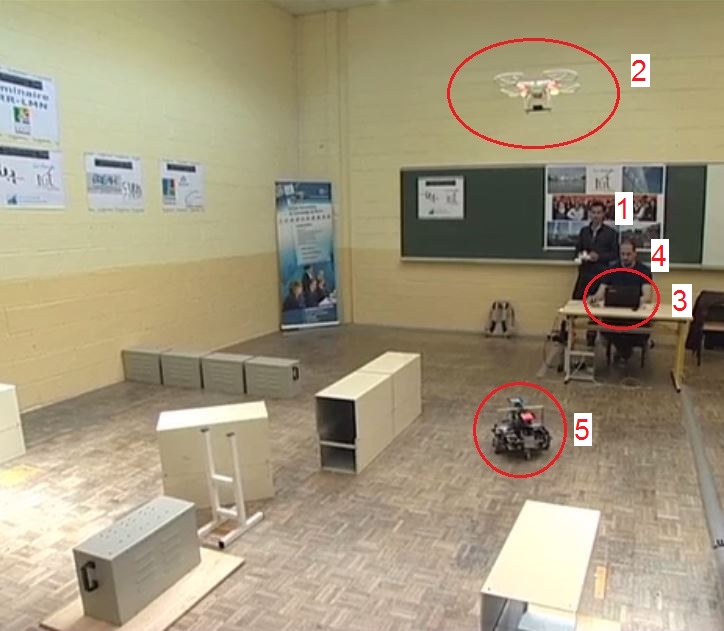} 
	\caption{Waypoint tracking experimentation.}
	\label{robot_drone} 
\end{figure}

Figure \ref{robot_drone} illustrate the UGV experimentation: The pilot (1) manipulates the UAV (2) that takes and send to the ground station a video flow (3). The operator (4) selects new waypoints to guide the UGV (5).

\begin{figure}[H] 
	\centering
	\includegraphics[width = 240px , height = 82px]{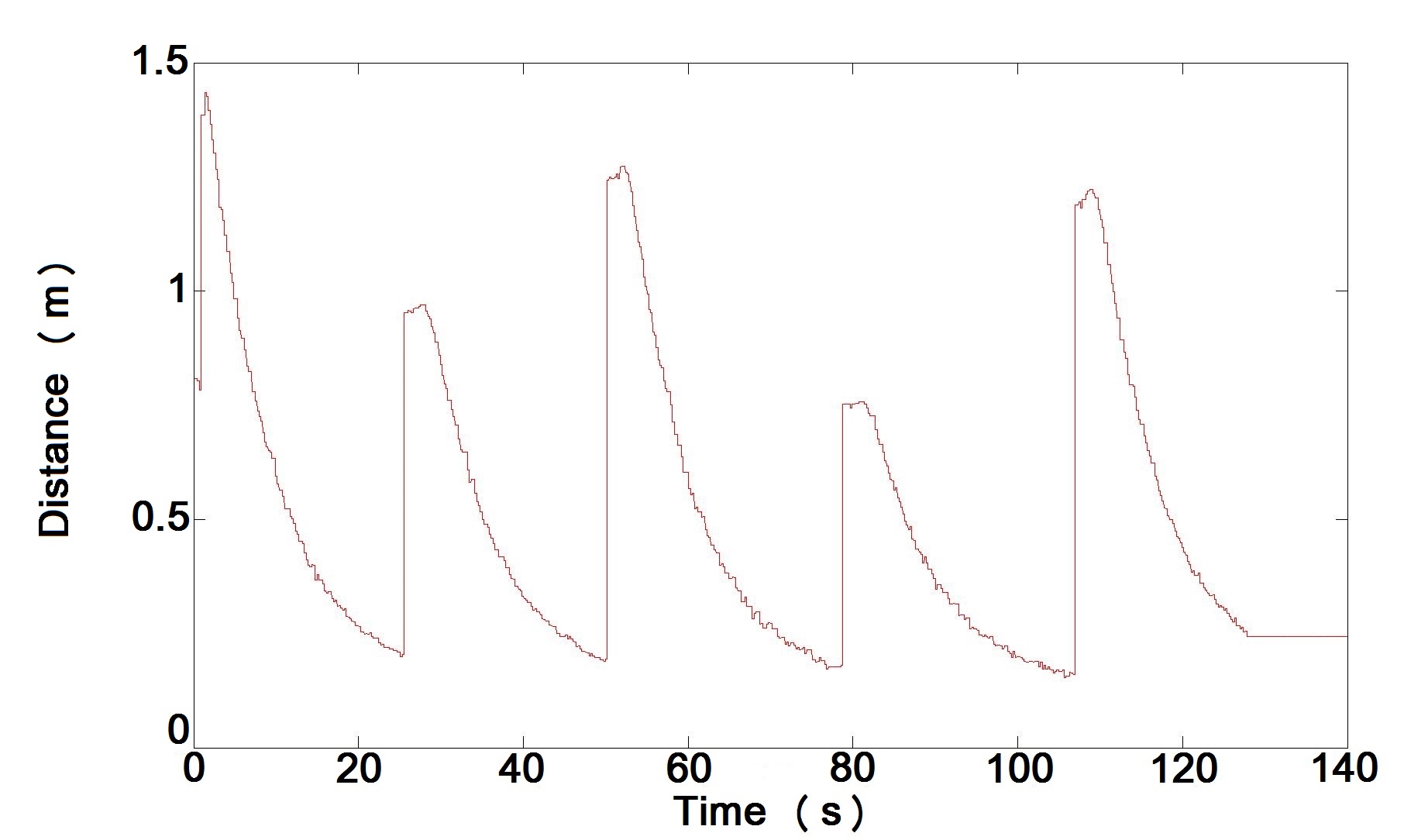} 
	\caption{Distance variation}
	\label{distance} 
\end{figure}

\begin{figure}[H] 
\centering
\includegraphics[width = 240px , height = 82px]{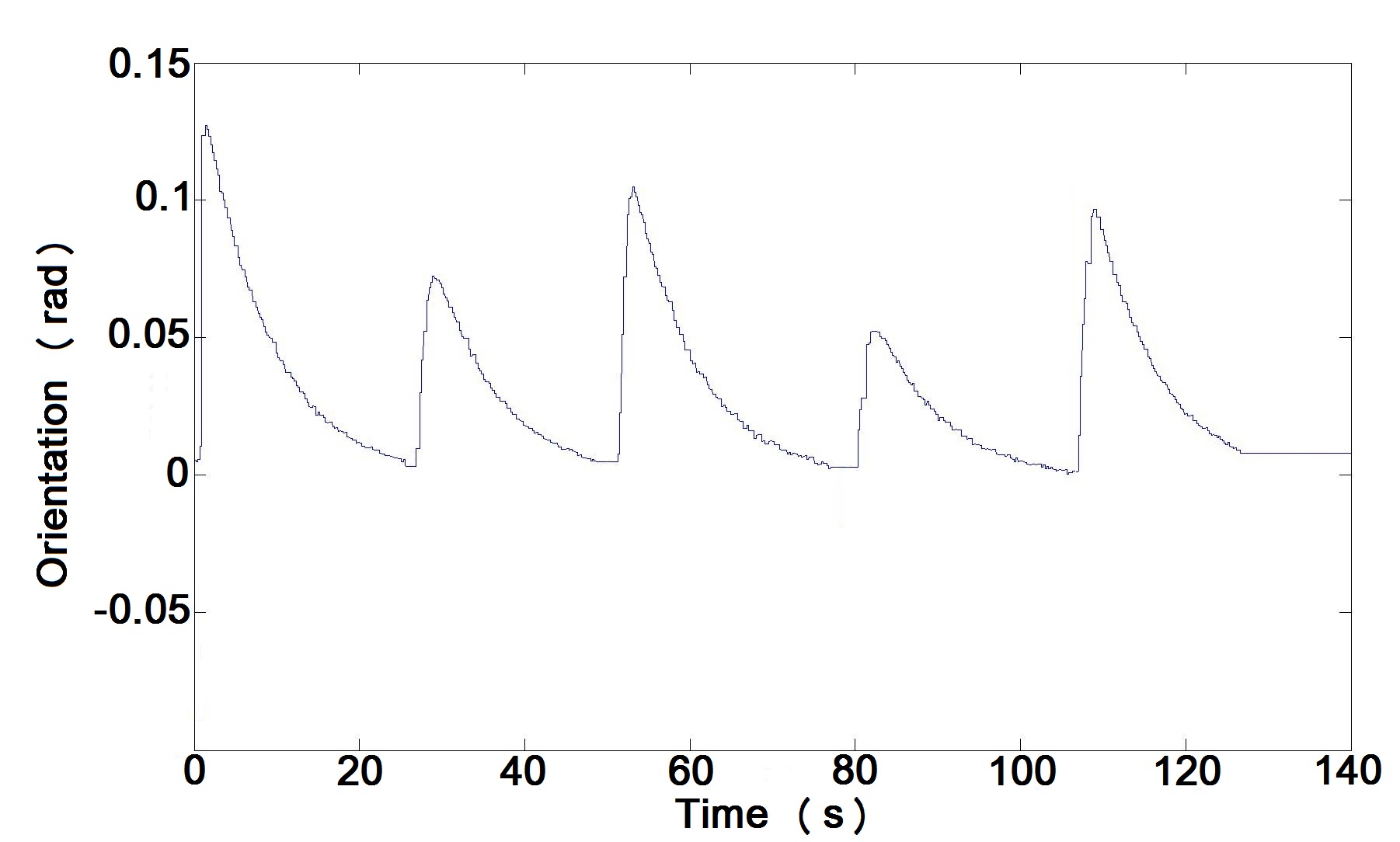}
\caption{Steering angle variation}
\label{orientation} 
\end{figure}

Figures \ref{distance} and \ref{orientation}, represent the distance \(d\) and the steering angle \(\alpha\) variations over time for a rectangular path selected by a human operator through the mission (send a new waypoint each time the UGV reaches the old one). Each peak on both graphs represent a new click by the user, which gives a new distance \(d\) and a new steering angle \(\alpha\). We can see that according to  (\ref{cmd_law_lead1}) the distance and the steering angle decrease exponentially and respectively to \(d\) = L = 0.15 m and \(\alpha\) = 0$^{\circ}$ which confirms the efficiency of our developed controller (\ref{cmd_law_lead}). 


\section{UAV results}\label{uav_res}
We present in this section only simulation results of the UAV following the UGV in its waypoints tracking because we do not yet have access to the UAV’s 
position measurements in the inertial frame (for example VICON MX system). That is why quantitative results are presented for the simulation trials only. Please refer to the links mentioned in section \ref{ugv_res} for experiments videos.

\subsection{Simulation result}
We used for simulation the parameters of an open-source quadcopter developed by \cite{lezconcept}. The parameters have been obtained thanks to experimentations fully described in \cite{harikmaster}:
\begin{table}[H]
	\centering
\begin{tabular}{|l|c|r|}
	\hline
	parameter & significance & values  \\
	\hline
	\(m\) & mass & 1.4 \(kg\) \\
	\(b\) & thrust coefficient & 1.3e-5 \(Ns^2\) \\
	\(d\) & drug coefficient & 1e-9 \(Nms^2\) \\
	\(J_r\) & rotor inertia & 6e-7 \(kg.m^2\) \\
	\(L\) & arm length & 1 \(m\) \\
	\(I_x\) & inertia on x axis & 0.02582 \(kg.m^2\)\\
	\(I_y\) & inertia on y axis & 0.02616 \(kg.m^2\)\\
	\(I_z\) & inertia on z axis & 0.04543 \(kg.m^2\)\\
	\(K_1\) & Positive gain 1 & 1.5 \\
	\(K_2\) & Positive gain 2 & 3 \\
	\hline
\end{tabular}
\caption{UAV parameters}
\end{table}

\begin{figure}[H]
	\centering 
	\includegraphics[width = 190px , height =180px]{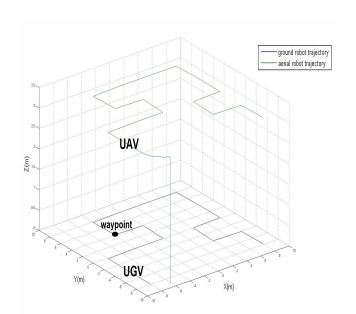} 
	\caption{Inspection scheme simulation}
	\label{simulation} 
\end{figure}

\begin{figure}[H]
	\centering 
	\includegraphics[width = 250px , height =200px]{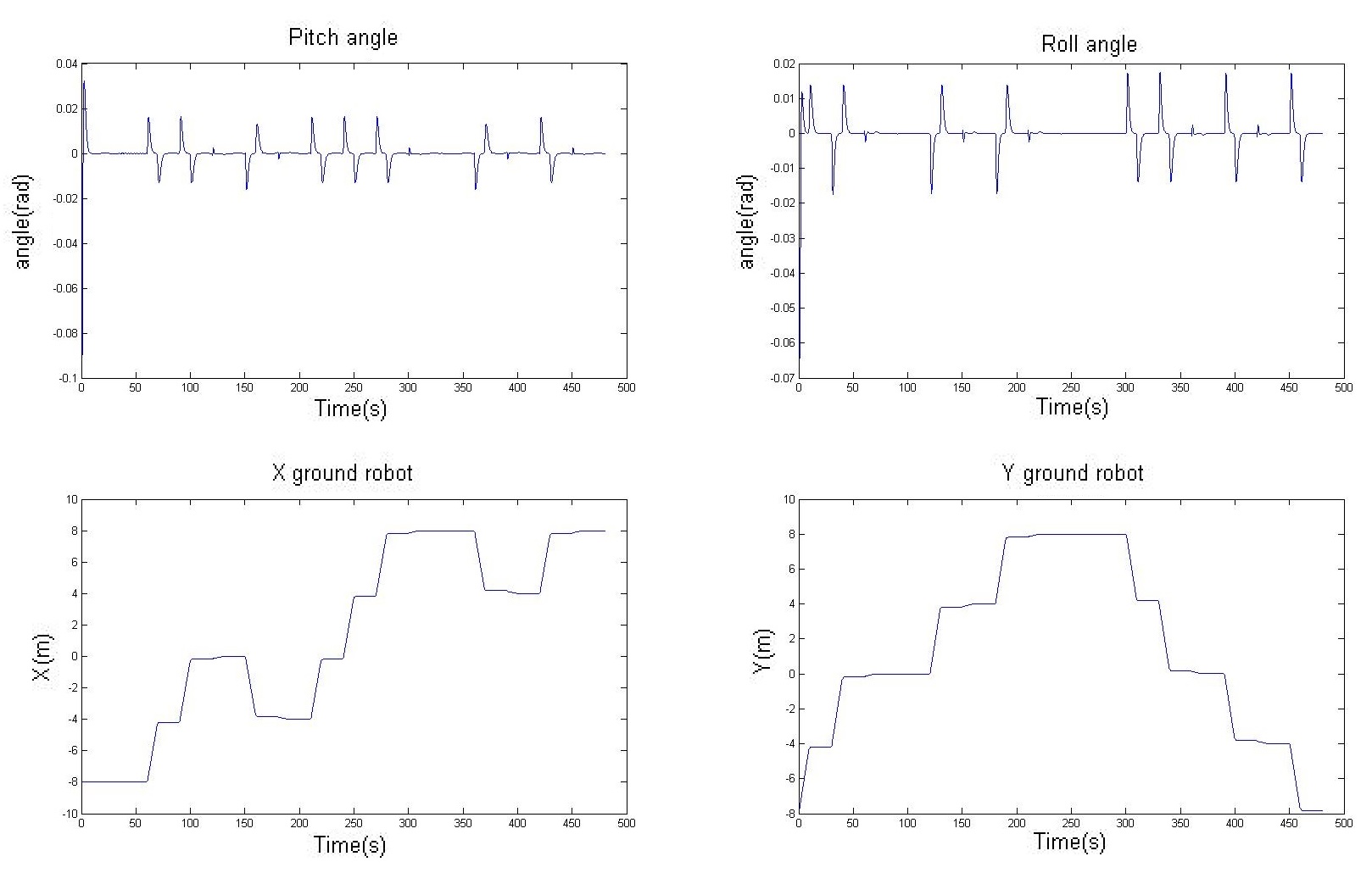} 
	\caption{UAV angles and UGV movement}
	\label{pitch_roll} 
\end{figure}

Figure \ref{simulation} shows that the UAV takes-off from an initial position (-6,-9) and flies to hover at 3m altitude, we suppose that at this height the UGV with an initial position (-8, -8) is in the coverage view of the UAV. We can see that the UGV moves followed by the UAV. Figures (\ref{pitch_roll}) show that the UAV changes its pitch and role angles according to the motion of the UGV.

\section{Conclusion and future work}\label{conclusion}
We presented in this paper a novel decentralized interactive architecture, where an air-ground-cooperation is deployed for area inspection. A UAV is used to provide a global coverage where a UGV is located to provide a local coverage. A human operator is introduced to provide waypoints for the UGV in order to safely navigate in the area. Our future work goes towards implementing an autonomous waypoints selection through free-space navigation extraction, which allow the testbed to be extended in more than one UGV, and let the human operator focus on other coordination tasks.  
\section{Acknowledgments}	
The authors would like to thank Le Havre Town Council CODAH for their support under research grants, and "DroneXTR" for piloting the drone. They would like as well to thank Upper Normandie Region through their support to PCMAI project.  	
	
	%----------------------------------------------------------------------------------------
	%	REFERENCE LIST
	%----------------------------------------------------------------------------------------
	
	\bibliography{biblio_ICCAR_2015_V0}
	\bibliographystyle{unsrt}
	
	%----------------------------------------------------------------------------------------
	
\end{document}